\documentclass[10pt,conference]{IEEEtran}
\IEEEoverridecommandlockouts
\usepackage{cite}
\usepackage{amsmath,amssymb,amsfonts}
\usepackage{algorithmic}
\usepackage{graphicx}
\usepackage{textcomp}
\usepackage{xcolor}
\usepackage{booktabs} 
\usepackage{multirow}
\usepackage{subcaption}
\usepackage{float}
\usepackage{url}
\usepackage{makecell}

\def\BibTeX{{\rm B\kern-.05em{\sc i\kern-.025em b}\kern-.08em
    T\kern-.1667em\lower.7ex\hbox{E}\kern-.125emX}}
\begin{document}

\title{Process Rewards for Outcome-Guided Reasoning Steps in LLM Reinforcement Learning}
%Safe Process Reward Integration for Mathematical Reasoning with Reinforcement Learning\\
% {\footnotesize \textsuperscript{*}Note: Sub-titles are not captured for https://ieeexplore.ieee.org  and
% should not be used}
% \author{\IEEEauthorblockN{Anonymous Authors}}

\author{
\IEEEauthorblockN{1\textsuperscript{st} Mohammad Rezaei Ravari}
\IEEEauthorblockA{\textit{Faculty of Computer Science} \\
\textit{Technical University of Dresden}\\
Dresden, Germany \\
mohammad.rezaei@tu-dresden.de}
\and
\IEEEauthorblockN{2\textsuperscript{nd} Jens Lehmann\IEEEauthorrefmark{1}}
\IEEEauthorblockA{\textit{Amazon Science} \\
\textit{Technical University of Dresden}\\
Dresden, Germany \\
jens.lehmann@tu-dresden.de}
\and
\IEEEauthorblockN{3\textsuperscript{rd} Sahar Vahdati}
\IEEEauthorblockA{\textit{TIB Leibniz Information Centre} \\
\textit{Leibniz University of Hanover}\\
Hanover, Germany \\
sahar.vahdati@tib.eu}
\thanks{\IEEEauthorrefmark{1}This work was done outside of Amazon.}
}

\maketitle

\begin{abstract}
Mathematical reasoning in large language models has improved substantially with reinforcement learning using verifiable rewards, where final answers can be checked automatically and converted into reliable training signals. 
Most such pipelines optimize outcome correctness only, which yields sparse feedback for long, multi-step solutions and offers limited guidance on intermediate reasoning errors. 
Recent work therefore introduces process reward models (PRMs) to score intermediate steps and provide denser supervision. 
In practice, PRM scores are often imperfectly aligned with final correctness and can reward locally fluent reasoning that still ends in an incorrect answer. When optimized as absolute rewards, such signals can amplify fluent failure modes, induce reward hacking, and destabilize policy updates. 
Existing PRM-based methods improve PRM quality, filter trajectories, or modify reward aggregation, but they do not directly constrain how process rewards interact with outcome correctness during optimization.
We propose \textbf{PROGRS}, a framework that leverages PRMs while keeping outcome correctness dominant. 
PROGRS treats process rewards as \emph{relative preferences within outcome groups} rather than absolute targets. 
We introduce \emph{outcome-conditioned centering}, which shifts PRM scores of incorrect trajectories to have zero mean within each prompt group. 
This removes systematic bias while preserving informative relative rankings.
To stabilize process signals, PROGRS combines a frozen \emph{quantile-regression PRM} with a \emph{multi-scale coherence evaluator} that penalizes short-window confidence volatility. We integrate the resulting centered process bonus into \emph{Group Relative Policy Optimization (GRPO)} without auxiliary objectives or additional trainable components. 
Across MATH-500, AMC, AIME, MinervaMath, and OlympiadBench, PROGRS consistently improves Pass@1 over outcome-only baselines (e.g., $74.9\%$ vs.\ $69.7\%$ on MATH-500; $59.0\%$ vs.\ $52.0\%$ on AMC-2023) and achieves stronger performance with fewer rollouts. 
These results show that outcome-conditioned centering enables safe and effective use of process rewards for mathematical reasoning.
\end{abstract}


\section{Introduction}
Large language models (LLMs) achieve strong performance on many natural language tasks, but complex, multi-step mathematical reasoning remains challenging.
Models often produce hallucinated steps, brittle chains of thought, or overconfident yet invalid solutions \cite{lightman2023lets,guo2025deepseek}.
Mathematical reasoning requires precise logical dependency tracking, structured decomposition, and verifiable solution construction.

A widely adopted direction is reinforcement learning (RL) with LLMs for mathematical reasoning \cite{guo2025deepseek,shao2024deepseekmath}.
RL with verifiable rewards (RLVR) leverages automatically checkable correctness signals to optimize final-answer accuracy \cite{lightman2023lets,wang2025rlvr}.
Most RLVR pipelines still rely on sparse outcome-only reward models, which provide limited guidance for long reasoning trajectories.
Process reward models (PRMs) were introduced to provide denser supervision.
PRMs assign scores to intermediate reasoning steps, so optimization can reflect trajectory quality rather than only the final outcome \cite{lightman2023lets,wang2025towards}.
In practice, PRMs are often miscalibrated and can assign high scores to locally coherent reasoning that still ends in an incorrect answer.
Using PRM scores naively as absolute rewards can induce reward hacking and destabilize training \cite{skalse2022defining,pan2022effects,miao2024inform}.

Recent methods mitigate these issues through generative judges with thought-level evaluation and capability-adaptive rewards \cite{he2025tp}, theoretical derivations from entropy-regularized RL \cite{yao2026prl}, or sample filtering based on score consistency \cite{ye2025prof}.
These strategies improve evaluation, grounding, or data reliability.
A key gap remains.
Existing approaches do not explicitly control how process rewards interact with outcome supervision during optimization.

We propose \textbf{Process-Reward Outcome-Guided Reasoning Steps (PROGRS)}
\footnote{https://github.com/NIMI-research/PROGRS-Process-Reward}, a framework integrating process-level guidance into RLVR while preserving outcome dominance.
%PROGRS is built around a single principle.
In PROGRS, process rewards act as \emph{relative preferences} within groups defined by outcome quality.
%, not as absolute optimization targets.
To enforce this principle, we introduce \emph{outcome-conditioned centering}, which removes systematic positive bias from PRM scores on incorrect trajectories while retaining informative relative rankings.
PROGRS constructs stable process signals using three components.
First, it uses calibrated step-level PRMs \cite{park2025know}.
Second, it applies a hierarchical multi-scale coherence evaluator that aggregates PRM scores over short reasoning windows and penalizes abrupt confidence fluctuations.
Third, it forms trajectory-level process scores that are centered on final outcomes and combined with outcome-based advantages during optimization.
PROGRS introduces no new trainable components and operates within standard Group Relative Policy Optimization (GRPO) \cite{shao2024deepseekmath}, ensuring correctness remains the dominant signal.

%Compared to prior PRM-based integration strategies, PROGRS constrains process rewards through outcome-conditioned centering while leveraging hierarchical coherence to extract stable, relative signals from noisy step-level estimates.
% Our contributions are:
% (1) identifying outcome-conditioned centering as a practical mechanism to safely integrate PRMs into RLVR,
% (2) introducing a hierarchical coherence evaluator that captures local reasoning instability from PRM score dynamics, and
% (3) demonstrating that combining these components within GRPO improves performance on benchmarks including MATH-500, AMC, AIME, and MinervaMath.

% ---------------------- RELATED WORK ------------------------
\section{Related Work}
\label{sec:related}
We review relevant strands of prior work: RL for mathematical reasoning with verifiable rewards, process reward models, and efforts to reconcile process and outcome signals in RLVR.
%RLVR has become a standard way to improve mathematical reasoning in LLMs by optimizing final-answer correctness.
%However, outcome-only rewards are sparse for long solutions, motivating process-level supervision through Process Reward Models (PRMs). A persistent obstacle is that PRM scores are not perfectly aligned with outcome correctness.
%They can reward locally plausible reasoning that still produces an incorrect final answer. Our work focuses on this interaction point and questions how process-level signals should enter policy gradients \emph{while preserving outcome correctness as the dominant optimization target}.

\textbf{RL for Mathematical Reasoning with Verifiable Rewards.}
RLVR optimizes policies using automatically checkable outcome rewards, typically binary correctness for final answers~\cite{shao2024deepseekmath,guo2025deepseek,wang2025towards}. 
Because outcome feedback is sparse, stable policy optimization is critical. 
Classic policy-gradient methods (e.g., REINFORCE) and variance-reduction techniques (e.g., GAE) underpin modern RL pipelines, while trust-region style objectives (TRPO, PPO) stabilize updates and are widely used in RLHF settings~\cite{williams1992reinforce,schulman2015high,schulman2015trust,schulman2017ppo,ouyang2022instructgpt,bai2022constitutional,rafailov2023direct}.
For reasoning tasks, GRPO~\cite{shao2024deepseekmath,guo2025deepseek} improves stability by normalizing advantages within groups of sampled solutions per prompt. DAPO~\cite{yu2025dapo} further stabilizes updates via asymmetric clipping. 
We build on GRPO-style training and use asymmetric clipping as an orthogonal stabilization method.
Our main contribution is a modified advantage that controls how process signals interact with outcome correctness.

\textbf{Process Reward Models (Supervision, Calibration and Failure Models).} 
PRMs provide denser feedback than outcome-only reward models (ORMs)~\cite{lightman2023lets,wang2025towards}. Early PRMs relied on human annotations~\cite{lightman2023lets}, while later works reduced labeling cost through automated rollouts~\cite{yang2025beyond,wang2025towards}, adaptive search labeling~\cite{sun2025efficient}, and LLM-based or hybrid pipelines~\cite{yang2024qwen2}. 
However, due to miscalibration, PRMs may assign high scores to trajectories that are coherent locally yet incorrect globally. 
Optimizing such scores as absolute rewards can induce reward hacking and amplify fluent failure modes~\cite{skalse2022defining,pan2022effects,miao2024inform}.
Several approaches mitigate this issue by down-weighting rewards or aggregating step signals independently, but they do not constrain process feedback with outcome supervision during gradient computation.
Recent calibrated PRMs~\cite{park2025know} estimate step-level success probabilities and uncertainty (e.g., via quantile heads), enabling uncertainty-aware selection at inference time. 
In our setting, a calibrated PRM is a fixed training-time evaluator with a focus on how to use imperfect process scores without overriding outcome correctness during optimization.

\textbf{Reconciling Process and Outcome Signals in RLVR.}

Recent methods integrate process supervision into RLVR in different ways.
TP-GRPO~\cite{he2025tp} addresses step segmentation ambiguity with thought-level evaluation and introduces capability-adaptive rewards to reduce repeated-step reward exploitation. PRL~\cite{yao2026prl} derives process rewards from entropy-regularized RL objectives, avoiding expensive search-based labeling. PROF~\cite{ye2025prof} reduces PRM--ORM conflicts by filtering trajectories based on consistency between process and outcome evaluations. These approaches improve PRM quality, grounding, or data reliability, and are complementary
to our goal.
Our focus is the missing \emph{semantic constraint} in PRM-based RLVR pipelines. Even with better PRMs or filtering, positive process bonuses on incorrect trajectories can distort optimization. We address this by centering within outcome groups only for incorrect trajectories, preserving outcome dominance.
% ---------------------- PROPOSED METHOD  --------------------
\section{Proposed Method}
\label{sec:method}

\textbf{PROGRS} is an advantage construction method for reinforcement learning with verifiable rewards in mathematical reasoning. 
It operates over prompt-level groups of sampled solutions and combines an outcome-based advantage with a \emph{centered} PRM-derived process bonus. 
This makes process scores act as relative preferences among incorrect trajectories, preventing fluent but incorrect reasoning from being systematically favored while preserving outcome correctness as the dominant signal.

\subsection{Advantage Construction}
\label{subsec:advantage}

\noindent\textbf{Group sampling and signals.}
For each prompt $q$, we sample a group of $K$ trajectories $\{o^{(i)}\}_{i=1}^K$ from the policy.
Each trajectory receives:
(i) an outcome reward $r_{\text{outcome}}^{(i)}\in\{0,1\}$ indicating final-answer correctness, and
(ii) a trajectory-level process score $S_{\text{PRM}}^{(i)}$ computed in
Section~\ref{subsec:process_score}.
\subsubsection{Outcome-Based Advantage}
\label{subsec:outcome_adv}
We compute a normalized outcome-based advantage within each prompt group:
\begin{equation}
\label{eq:base_advantage}
A_{\text{outcome}}^{(i)}=
\begin{cases}
\dfrac{r_{\text{outcome}}^{(i)}-\bar r_{\text{outcome}}}{\sigma_{\text{outcome}}+\epsilon},
& \sigma_{\text{outcome}}\ge \epsilon,\\[6pt]
0, & \text{otherwise},
\end{cases}
\end{equation}
where $\bar r_{\text{outcome}}$ and $\sigma_{\text{outcome}}$ are computed over the $K$ samples
within the same prompt group, and $\epsilon>0$ is a small constant.

\subsubsection{Outcome-Conditioned Centering}
\label{subsec:centering}
PRMs can assign high process scores to incorrect but locally fluent solutions. If used as an
absolute bonus, this can systematically reinforce incorrect trajectories.
PROGRS removes this offset by centering process scores \emph{within the incorrect subset}.

Let $\mathcal{I}=\{i \mid r_{\text{outcome}}^{(i)}=0\}$ denote the indices of incorrect samples in the group and define
\[
\mu_{\text{PRM}}^{\text{incorrect}}=
\begin{cases}
\dfrac{1}{|\mathcal{I}|}\sum_{i\in\mathcal{I}} S_{\text{PRM}}^{(i)}, & |\mathcal{I}|>0,\\[6pt]
0, & |\mathcal{I}|=0.
\end{cases}
\]
We then center:
\begin{equation}
\label{eq:trajectory_score_centered}
\tilde{S}_{\text{PRM}}^{(i)}=
\begin{cases}
S_{\text{PRM}}^{(i)}, & r_{\text{outcome}}^{(i)}=1,\\
S_{\text{PRM}}^{(i)}-\mu_{\text{PRM}}^{\text{incorrect}}, & r_{\text{outcome}}^{(i)}=0.
\end{cases}
\end{equation}
We apply centering only to incorrect trajectories. Applying it to correct trajectories would reduce separation between correct and incorrect solutions, weakening outcome dominance, the primary optimization objective in RLVR. In preliminary experiments, this reduced performance.
By construction, the mean of $\tilde{S}_{\text{PRM}}^{(i)}$ over $i\in\mathcal{I}$ is zero, so incorrect samples receive no systematic process bonus.

\subsubsection{Final Advantage Integration}
\label{subsec:final_adv}
We combine outcome and centered process signals additively:
\begin{equation}
\label{eq:final_advantage}
A_{\text{final}}^{(i)} = A_{\text{outcome}}^{(i)} + \lambda_{\text{PRM}}\,\tilde{S}_{\text{PRM}}^{(i)},
\end{equation}
where $\lambda_{\text{PRM}}$ controls the strength of the process guidance.
When all $K$ samples in a prompt group share the same outcome, $A_{\text{outcome}}^{(i)}=0$ and learning is driven by
within-group preferences induced by $\tilde{S}_{\text{PRM}}^{(i)}$.

\subsection{Computing the Process Score $S_{\text{PRM}}$}
\label{subsec:process_score}

We compute $S_{\text{PRM}}$ from step-level PRM scores and a windowed coherence penalty.

\subsubsection{Frozen Quantile-Regression PRM for Step Scoring}
\label{subsec:calibrated_prm}

We use a frozen quantile-regression Process Reward Model (PRM) from~\cite{park2025know} as an external evaluator,
using the released checkpoint without additional fine-tuning.

\noindent\textbf{Step delimitation and representation extraction.}
We represent each reasoning trajectory as a sequence of $T$ textual steps $s_{1:T}$
and insert a dedicated delimiter token \texttt{<extra\_0>} immediately after each step.
We define the full token sequence
\[
x=\mathrm{concat}\big(q,\, s_1,\, \texttt{<extra\_0>},\, \ldots,\, s_T,\, \texttt{<extra\_0>}\big).
\]
Running the decoder-only PRM with a causal attention mask yields hidden states at each delimiter position.
Let $h_t$ denote the hidden state at the delimiter immediately following step $s_t$.
Because the model is evaluated under a causal mask, $h_t$ depends only on tokens up to that delimiter,
i.e., only on the prefix $(q,s_{1:t})$.

\noindent\textbf{Step score.}
For each step representation $h_t$, the quantile heads predict success probability estimates
$q_\tau(h_t)$ for $\tau\in\{0.1,0.5,0.9\}$.
We use the median prediction as the step-level score: $r_{\text{fine}}(s_t)=q_{0.5}(h_t)$,
% \begin{equation}
% \label{eq:prm_score}
% r_{\text{fine}}(s_t)=q_{0.5}(h_t),
% \end{equation}
which serves as a robust point estimate of step-level correctness.

\noindent\textbf{Implementation note (prefix-only consistency).}
Conceptually, step $t$ can be scored by a prefix-only PRM call on
$x_t=\mathrm{concat}(q,s_1,\texttt{<extra\_0>},\ldots,s_t,\texttt{<extra\_0>})$.
In practice, extracting delimiter states from a single forward pass on $x$ is equivalent under a causal mask.
\subsubsection{Windowed Coherence and Trajectory Aggregation}
\label{subsec:hierarchical}

While step-level PRM scores capture local confidence, they do not reflect the stability of reasoning across neighboring steps.
Across all benchmarks, reasoning trajectories are relatively short, typically ranging between 12 and 18 steps on average. This consistent scale motivates the use of a small window size ($w=3$), which captures local coherence while maintaining sufficient granularity. In contrast, larger window sizes would reduce each trajectory to only a few segments, weakening the ability to detect local instability in reasoning.
We therefore compute a coherence-modulated process score using contiguous windows of step scores. We use non-overlapping windows to reduce correlation between adjacent estimates and to avoid redundant smoothing effects observed with overlapping windows. In practice, overlapping windows introduced highly correlated scores that provided limited additional signal while increasing computational cost.

\noindent\textbf{Windowed variance analysis.}
Given $\{r_{\text{fine}}(s_t)\}_{t=1}^{T}$, we partition the trajectory into contiguous, non-overlapping windows of nominal size $w$.
Let $N_w=\lceil T/w\rceil$ and define
\[
W_j=\{(j-1)w+1,\ldots,\min(jw,T)\},\quad j\in\{1,\ldots,N_w\}.
\]
% For each window $j$, we compute the local mean and standard deviation:
% % \[
% % \mu_j=\frac{1}{|W_j|}\sum_{t\in W_j} r_{\text{fine}}(s_t),\qquad
% % \sigma_j=\sqrt{\frac{1}{|W_j|}\sum_{t\in W_j}\big(r_{\text{fine}}(s_t)-\mu_j\big)^2 }.
% % \]
% \begin{align}
% \mu_j &= \frac{1}{|W_j|}\sum_{t\in W_j} r_{\text{fine}}(s_t),\\
% \sigma_j &= \sqrt{\frac{1}{|W_j|}\sum_{t\in W_j}\big(r_{\text{fine}}(s_t)-\mu_j\big)^2 }.
% \end{align}
% If $|W_j|=1$, we set $\sigma_j=0$.
For each window \(j\), we compute the local mean and standard deviation, under setting of \(\sigma_j=0\) when \(|W_j|=1\):
\begin{align}
\mu_j &= \frac{1}{|W_j|}\sum_{t\in W_j} r_{\text{fine}}(s_t), &
\sigma_j &= \sqrt{\frac{1}{|W_j|}\sum_{t\in W_j}\big(r_{\text{fine}}(s_t)-\mu_j\big)^2 }.
\end{align}

\noindent\textbf{Coherence-modulated window score.}
To translate local variability into a coherence signal, we define
\begin{equation}
\label{eq:coherence}
r_{\text{coh},j}(w)=
\mu_j \cdot \exp\!\left(-\lambda_{\text{var}}\frac{\sigma_j}{\mu_j+\epsilon}\right),
\end{equation}
where $\lambda_{\text{var}}>0$ controls the penalty strength and $\epsilon>0$ ensures numerical stability.
This yields a multiplicative downweighting of high-variance windows without imposing hard thresholds.

\noindent\textbf{Hierarchical aggregation with coherence blending.}
To balance raw step quality with local stability, we compute a blended window score:
\begin{equation}
\label{eq:hierarchical_score}
R_j(w)=\alpha_{\text{coh}}\cdot r_{\text{coh},j}(w) + (1-\alpha_{\text{coh}})\cdot \mu_j,
\end{equation}
and aggregate across windows to obtain a trajectory score at scale $w$:
\[
S_{\text{PRM}}(w)=\frac{1}{N_w}\sum_{j=1}^{N_w} R_j(w).
\]
\noindent\textbf{Multi-scale (optional) form.}
More generally, one can average across window sizes $\mathcal{W}$:
\[
S_{\text{PRM}}=\frac{1}{|\mathcal{W}|}\sum_{w\in\mathcal{W}} S_{\text{PRM}}(w).
\]
In our main experiments we use a single scale $w=3$ (i.e., $\mathcal{W}=\{3\}$), with
$\lambda_{\text{var}}=2.0$ and $\alpha_{\text{coh}}=0.6$.

\subsection{Policy Optimization with GRPO and Asymmetric Clipping}
\label{subsec:optimization}

We optimize with Group Relative Policy Optimization (GRPO) and asymmetric clipping following DAPO~\cite{yu2025dapo},
using $A_{\text{final}}^{(i)}$ from Eq.~\eqref{eq:final_advantage}:
\begin{equation}
\label{eq:policy_loss}
\mathcal{L}_{\text{policy}}=
-\frac{1}{B}\sum_{i=1}^{B}
\min\!\left(
r_i A_{\text{final}}^{(i)},
\mathrm{clip}_{\epsilon_l,\epsilon_h}(r_i)\, A_{\text{final}}^{(i)}
\right),
\end{equation}
where $B$ is the number of sampled trajectories in the minibatch (across prompts and samples) and
$r_i=\pi_\theta(o^{(i)}\mid q)/\pi_{\text{ref}}(o^{(i)}\mid q)$ is the sequence-probability ratio
between the current policy and a reference policy.

Heuristic though it is, this construction admits a control-variate interpretation, suggesting a variance-reduction effect.

\subsection{Why Outcome-Conditioned Centering Improves Stability}
\label{sec:centering_analysis_main}

Outcome-conditioned centering addresses a specific failure mode of PRM-based
training: an incorrect trajectory can receive a high process score because its
intermediate reasoning is locally coherent, even when the final answer is
wrong. Consider a representative case in which the model solves the problem of
maximizing
$\frac{wx+xy+yz}{w^2+x^2+y^2+z^2}$ for positive real numbers $w,x,y,z$.
The correct optimum is $\frac{1+\sqrt{5}}{4}$, attained when $w=z$, $x=y$,
and $\frac{w}{x}=\frac{\sqrt{5}-1}{2}$. However, the model instead assumes
$w=x=y=z$ and predicts $\frac{3}{4}$. This solution is globally incorrect, but
its reasoning is well structured and applies plausible inequalities, derives
consistent equality conditions, and exhibits low local variance in step-level
PRM scores. As a result, the raw trajectory-level process score remains high.
If such scores are used directly, optimization may systematically favor fluent
but incorrect trajectories.

Outcome-conditioned centering prevents this by converting the PRM signal on
incorrect trajectories from an absolute reward into a relative preference. In
the above example, the raw process score is $S_{\text{PRM}}=0.736$, while the
mean PRM score over incorrect samples in the same prompt group is $0.643$.
Centering therefore yields
$\tilde{S}_{\text{PRM}} = S_{\text{PRM}} - \mu_{\text{PRM}}^{\text{incorrect}}
= 0.093$,
so the trajectory receives only a small positive bonus relative to other
incorrect samples, rather than a large absolute reward. This preserves
intra-group ranking among incorrect solutions while removing the systematic
positive offset that would otherwise compete with outcome supervision.

Formally, let $\mathcal{I}=\{i \mid r_{\text{outcome}}^{(i)}=0\}$ be the set
of incorrect trajectories and $z_i := \nabla_\theta \log \pi_\theta(o^{(i)} \mid q)$
the score function. The PRM contribution to the gradient is proportional to
$\sum_{i\in\mathcal{I}} S_{\text{PRM}}^{(i)} z_i$, while outcome-conditioned
centering replaces $S_{\text{PRM}}^{(i)}$ by
$S_{\text{PRM}}^{(i)} - \mu_{\text{PRM}}^{\text{incorrect}}$ and therefore
subtracts the term
$\lambda_{\text{PRM}} \mu_{\text{PRM}}^{\text{incorrect}} \sum_{i\in\mathcal{I}} z_i$.
By the score-function identity, the expectation of $\sum_{i\in\mathcal{I}} z_i$
(conditioned on the prompt) is zero, so removing this rank-one common-mode
component does not change the expected gradient but reduces its variance. In
this sense, outcome-conditioned centering acts as an unbiased control variate:
the PRM bonus within $\mathcal{I}$ has zero mean, and only relative deviations
around this baseline influence the update.

Consequently, incorrect trajectories cannot receive a net PRM uplift at the
group level, while correct trajectories remain separated by the outcome-based
advantage. This keeps outcome supervision in control of inter-group preference
learning and uses PRM scores primarily to refine intra-group ordering among
incorrect samples.

Finally, the validity of the PRM signal depends on prefix-causal step scoring.
We score step $t$ using the hidden state at the delimiter immediately following
$s_t$ in a decoder-only PRM with a causal attention mask, so the score depends
only on the prefix $(q,s_{1:t})$. In controlled tests, prefix-only evaluation
and full-trajectory extraction produced nearly identical scores, and perturbing
the suffix left the score unchanged up to numerical precision. This confirms
that the process signal used by \textsc{PROGRS} reflects prefix-consistent
reasoning quality rather than leakage from future tokens.
% % % ---------------------- Evaluations  --------------------
\section{Evaluations and Results}
\label{sec:experiments}
We evaluate PROGRS variants with 4 and 8 rollouts on six mathematical reasoning benchmarks, and against DAPO baselines using 8 and 16 rollouts.
We analyze accuracy, sample efficiency, computational cost, ablations, and
Pass@K.
\subsection{Experimental Setup}
\label{sec:exp_setup}

\begin{table*}[t]
\centering
\caption{Results and ablations across mathematical reasoning benchmarks. We report Pass@1 accuracy (upper block, in \%) and average generated tokens per problem (lower block; lower is better), both as mean $\pm$ standard deviation over ten independent runs. DAPO-$K$ denotes the outcome-only baseline trained with $K$ rollouts per prompt; PROGRS-$K$ denotes our method with the same rollout budget. Ablation rows remove individual components while keeping all other settings fixed: $\alpha_{\text{coh}}=0$ disables the coherence penalty (centering retained), and \textit{No Centering} removes outcome-conditioned centering (coherence retained).}
\label{tab:results_summary}
\small
\setlength{\tabcolsep}{3.5pt} % tighten horizontally (optional)
\renewcommand{\arraystretch}{1.05} % slightly tighter vertically (optional)
\resizebox{\textwidth}{!}{%
\begin{tabular}{c l llllll}
\toprule
\textbf{Metric} & \textbf{Model / Setting}
& \textbf{AIME 24}
& \textbf{AIME 25}
& \textbf{AMC 23}
& \textbf{MATH-500}
& \textbf{MinervaMath}
& \textbf{OlympiadBench} \\
\midrule

\multirow{6}{*}{\shortstack{\textbf{Pass@1}\\\textbf{(\%)}}}
& DAPO-16
& $8.67 \pm 4.00$
& $8.00 \pm 1.63$
& $52.00 \pm 3.67$
& $69.66 \pm 0.70$
& $18.79 \pm 1.47$
& $32.03 \pm 0.73$ \\

& DAPO-8
& $8.67 \pm 1.63$
& $\mathbf{12.33 \pm 2.13}$
& $48.75 \pm 5.62$
& $68.40 \pm 0.80$
& $14.63 \pm 0.59$
& $31.34 \pm 0.87$ \\

& PROGRS-4
& $10.67 \pm 2.00$
& $12.00 \pm 1.63$
& $50.25 \pm 2.61$
& $74.14 \pm 0.98$
& $\mathbf{23.64 \pm 0.66}$
& $\mathbf{35.74 \pm 0.89}$ \\

& PROGRS-8
& $\mathbf{12.00 \pm 3.06}$
& $10.67 \pm 2.91$
& $\mathbf{59.00 \pm 5.27}$
& $\mathbf{74.92 \pm 0.82}$
& $22.13 \pm 0.49$
& $35.06 \pm 0.61$ \\

% ---- separator before ablations (do not cut the multirow Metric cell) ----
\addlinespace[0.3ex]
\cmidrule(lr){2-8}
\addlinespace[0.2ex]

& \textit{Ablation:} $\alpha_{\text{coh}} = 0$
& $9.33 \pm 3.59$
& $9.67 \pm 2.33$
& $50.50 \pm 2.18$
& $71.82 \pm 1.07$
& $20.18 \pm 0.80$
& $33.52 \pm 0.58$ \\

& \textit{Ablation:} No Centering
& $12.00 \pm 3.06$
& $8.33 \pm 1.67$
& $54.00 \pm 3.74$
& $67.78 \pm 0.95$
& $20.15 \pm 0.75$
& $30.77 \pm 0.64$ \\

\midrule

\multirow{6}{*}{\shortstack{\textbf{Avg}\\\textbf{tokens}\\\textbf{($\downarrow$)}}}
& DAPO-16
& $1202.22 \pm 20.86$
& $1128.88 \pm 18.16$
& $873.95 \pm 12.23$
& $\mathbf{671.04 \pm 2.42}$
& $1110.73 \pm 4.48$
& $\mathbf{910.65 \pm 4.83}$ \\

& DAPO-8
& $1245.09 \pm 22.52$
& $1149.21 \pm 15.21$
& $\mathbf{831.90 \pm 19.82}$
& $673.86 \pm 2.21$
& $1168.00 \pm 3.03$
& $944.48 \pm 2.94$ \\

& PROGRS-4
& $1201.29 \pm 14.99$
& $\mathbf{1065.38 \pm 12.05}$
& $950.66 \pm 7.24$
& $694.70 \pm 2.62$
& $1081.73 \pm 5.61$
& $990.17 \pm 3.24$ \\

& PROGRS-8
& $\mathbf{1082.42 \pm 15.47}$
& $1156.79 \pm 15.73$
& $918.64 \pm 13.92$
& $683.74 \pm 2.67$
& $\mathbf{1043.08 \pm 4.99}$
& $958.38 \pm 3.53$ \\

% ---- separator before ablations (do not cut the multirow Metric cell) ----
\addlinespace[0.3ex]
\cmidrule(lr){2-8}
\addlinespace[0.2ex]

& \textit{Ablation:} $\alpha_{\text{coh}} = 0$
& $1018.21 \pm 30.21$
& $1027.08 \pm 11.72$
& $858.91 \pm 5.18$
& $669.70 \pm 3.20$
& $1012.28 \pm 3.37$
& $934.80 \pm 3.35$ \\

& \textit{Ablation:} No Centering
& $1147.96 \pm 21.58$
& $1273.70 \pm 17.86$
& $911.98 \pm 19.47$
& $699.68 \pm 2.10$
& $1016.18 \pm 3.70$
& $975.54 \pm 4.35$ \\

\bottomrule
\end{tabular}%
}
\end{table*}
\textbf{Training data and evaluation benchmarks.}
Following prior findings that harder examples provide stronger learning signals in RL fine-tuning~\cite{pikus2025hard}, we train all models on Level~5 problems from the Hendrycks MATH dataset~\cite{hendrycksmath2021}.
A preliminary study over three 2{,}000-sample subsets confirmed that Level~5
training yields the best downstream performance on MATH-500
(73.7\% Pass@1), outperforming Level~4 (73.01\%) and Level~3 (72.39\%).
This setup emphasizes cross-difficulty generalization while remaining
computationally feasible.
We evaluate on MATH-500~\cite{hendrycksmath2021},
MinervaMath~\cite{lewkowycz2022solving},
OlympiadBench~\cite{he2024olympiadbench},
AMC~2023, AIME 2024, and AIME 2025, covering both in-distribution and out-of-distribution mathematical reasoning.

\textbf{Models and reward functions.}
We fine-tune Qwen2.5-Math-1.5B-base~\cite{yang2024qwen2}.
Process-level signals are obtained from a frozen
Qwen2.5-Math-PRM-7B with quantile-regression heads from a recent work ~\cite{park2025know}.
We shall note that we used them without additional fine-tuning.
All performance gains therefore are the direct impact of PROGRS reward formulation.
We compare against outcome-only RLVR baselines using DAPO~\cite{yu2025dapo} with group sizes $B=8$ and $B=16$, under identical optimization settings.

\noindent We report Pass@1 accuracy using greedy decoding
(temperature $=0.0$, max length $=4096$).
Results are averaged over ten random seeds.
We also report the average number of generated tokens per problem as a measure of computational cost.

\subsection{Performance Comparison}
\label{sec:main_perf}

Table~\ref{tab:results_summary} reports Pass@1 accuracy (mean $\pm$ std over 10 runs) on six
mathematical reasoning benchmarks spanning in-distribution evaluation (MATH-500),
distribution-shifted settings (AMC/AIME), and harder problem sets (MinervaMath, OlympiadBench).
Overall, PROGRS improves accuracy over outcome-only baselines across most benchmarks while using
fewer training rollouts (PROGRS-4/8 vs.\ DAPO-8/16), indicating a more favorable accuracy--budget
trade-off.
Surprisingly, the results for AIME 24 in no-centering and PROGRS-8 are the same. However, PROGRS-8 achieves substantial token efficiency gains (1082$\pm$15 vs. 1148$\pm$22 tokens/problem) with improved numerical stability.

\noindent\textbf{Distribution shift vs.\ in-distribution.}
The clearest gains under shift appear on AMC-2023, where PROGRS-8 outperforms DAPO-16 (59.0\% vs.\ 52.0\%).
On AIME-2024/2025, absolute accuracies are low and variance is higher, but PROGRS-8 remains above DAPO-16. 
We interpret these as directional evidence rather than definitive improvements.
In contrast, on the in-distribution MATH-500 benchmark, gaps are smaller but consistent (e.g., PROGRS-8 at
74.9\% vs.\ DAPO-16 at 69.7\%), and PROGRS-4 matches or exceeds larger-rollout baselines, supporting improved
sample efficiency.

\noindent\textbf{Hardness and stability.}
On MinervaMath, PROGRS yields sizable gains over DAPO (e.g., PROGRS-4 at 23.6\% vs.\ DAPO-16 at 18.8\%),
consistent with the intuition that many problems admit structured partial progress even when the final answer
is wrong, making within-group process preferences useful. On OlympiadBench, improvements persist but narrow
(e.g., PROGRS-4 at 35.7\% vs.\ DAPO-16 at 32.0\%), suggesting diminishing returns from reward-structure
refinements near model capacity. Across several benchmarks, PROGRS also exhibits lower run-to-run variance,
consistent with outcome-conditioned centering reducing systematic PRM bias on incorrect trajectories and the
coherence aggregation reducing sensitivity to volatile step scores.

\subsection{Computational Efficiency}
\label{sec:efficiency}
\begin{figure}[t!]
\centering
\includegraphics[trim={2cm 0 2cm 0}, width=0.90\columnwidth]{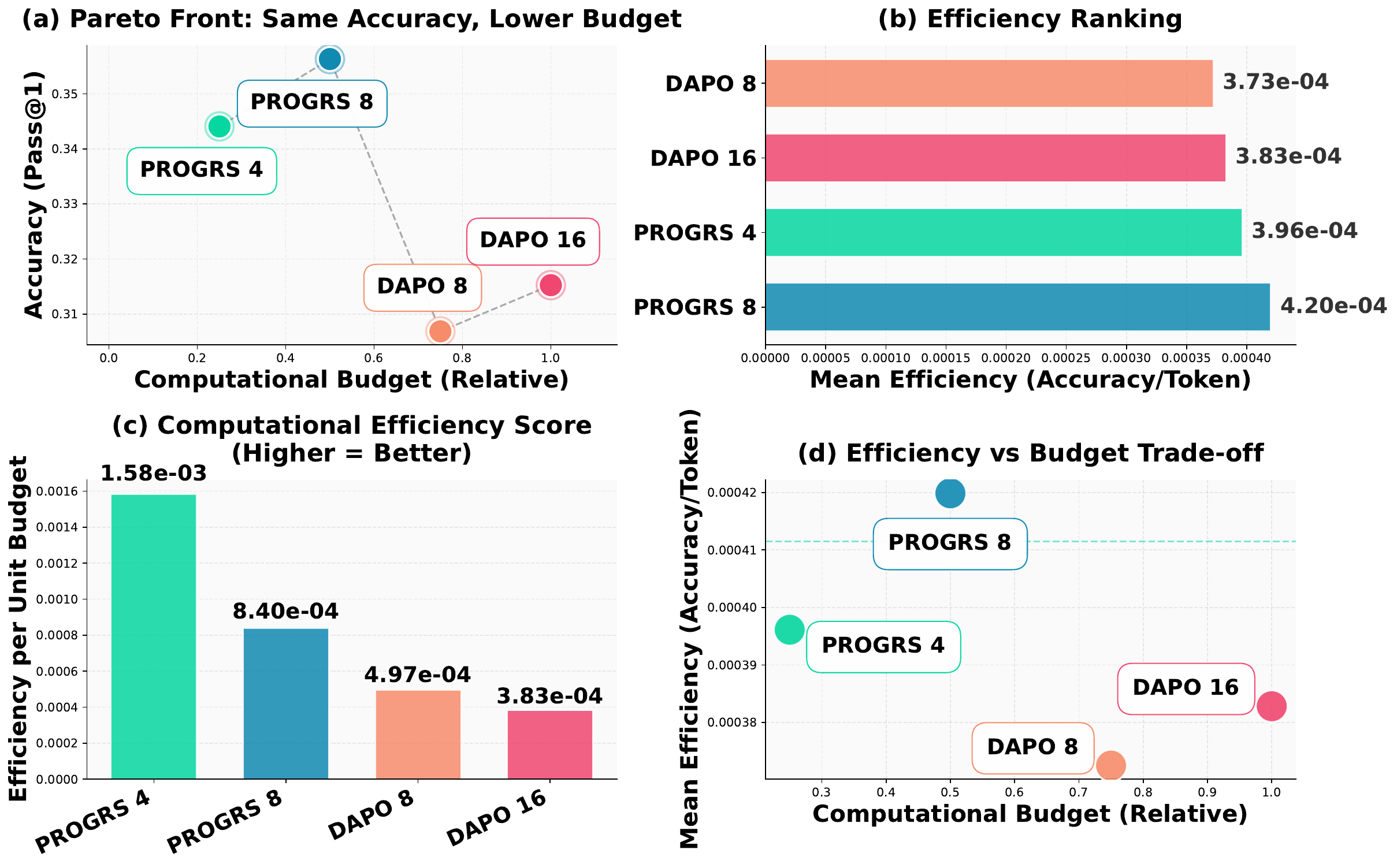}
\caption{Computational efficiency across models. 
(a) Pareto frontier: Accuracy vs. budget; PROGRS-4 matches DAPO baselines with $\sim50\%$ budget, PROGRS-8 matches/exceeds DAPO-16. 
(b) Efficiency ranking: Accuracy per token; PROGRS-8 highest, PROGRS-4 next. 
(c) Efficiency score: Accuracy normalized by budget; PROGRS-4 best. 
(d) Efficiency-budget surface: PROGRS-4 optimal for $\le 50\%$ budget, PROGRS-8 for $50\%-100\%$.}
\label{fig:efficiency}
\end{figure}

Our efficiency analysis focuses on rollout and token budgets and does not fully account for PRM inference costs. Since PRM scoring adds a forward pass at each reasoning step, training can be substantially more expensive than outcome-only RL. Thus, the reported gains reflect relative improvements under matched rollout settings, rather than absolute end-to-end compute efficiency.
We evaluate the accuracy--compute trade-off along two axes: \textbf{training-time rollout budget} (group size $K$)
and \textbf{inference-time generation length} (tokens per problem). Table~\ref{tab:results_summary} reports average
tokens per problem, and Fig.~\ref{fig:efficiency} summarizes accuracy relative to compute budgets.

\noindent\textbf{Rollout-budget efficiency.}
PROGRS is designed to improve the learning signal per rollout by using centered process preferences when outcome
advantages are uninformative. Empirically, PROGRS-4 matches or exceeds DAPO baselines that use substantially larger
rollout budgets (e.g., PROGRS-4 vs.\ DAPO-16 on MATH-500), and PROGRS-8 is generally competitive with or better than
DAPO-16, indicating a more favorable accuracy--budget frontier.

\noindent\textbf{Token usage.}
Token counts vary by benchmark (Table~\ref{tab:results_summary}). PROGRS does not uniformly increase generation
length: it is shorter on some datasets and longer on others. Taken together with the Pass@1 improvements, this
suggests gains are not explained solely by producing longer solutions, but by changes in solution quality induced
by the training objective.

\begin{figure}
    \centering
    \includegraphics[width=\linewidth]{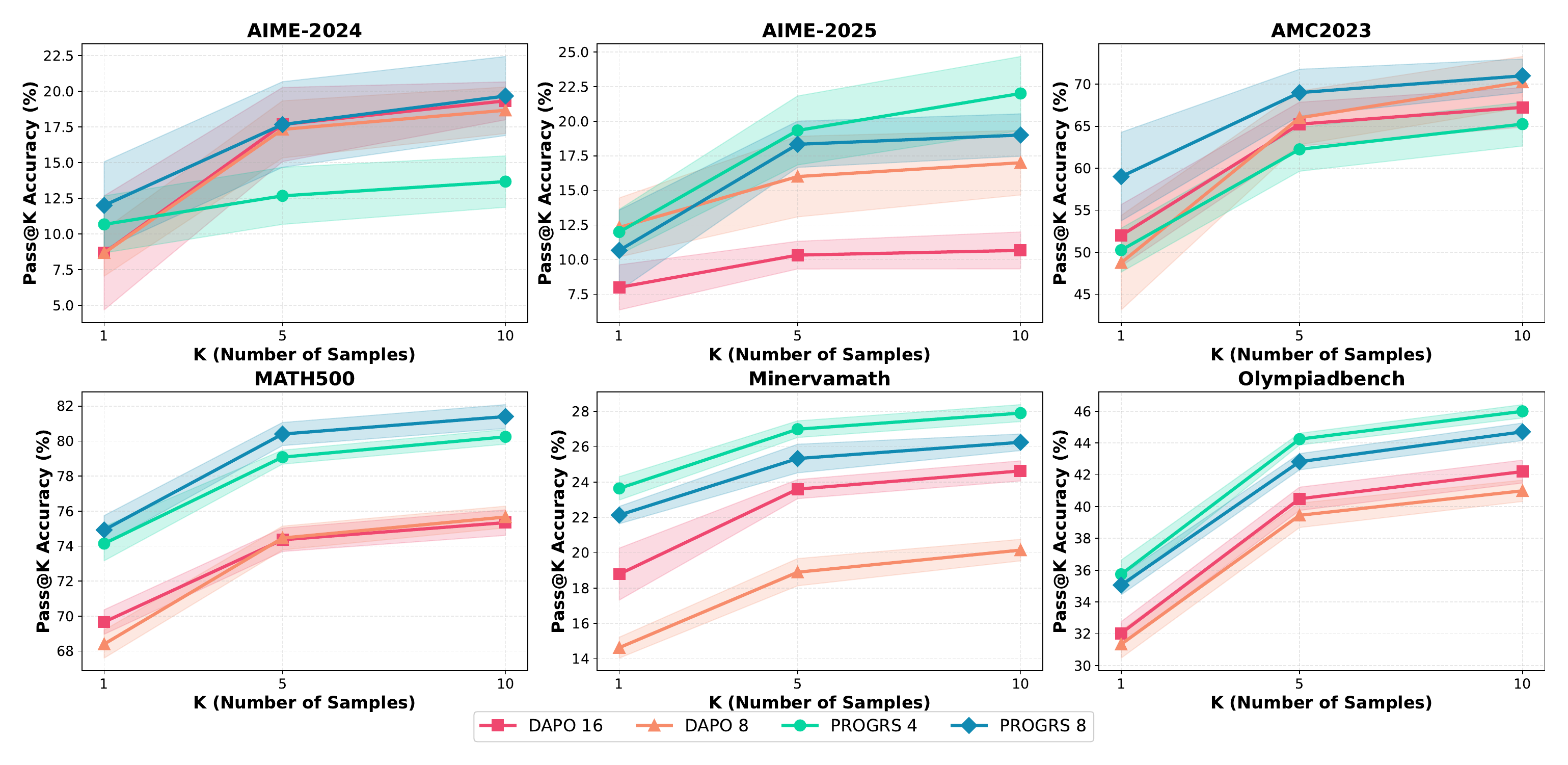}
    \caption{Pass@K accuracy across benchmarks. Each panel shows Pass@{1, 5, 10} for DAPO-16 (red), DAPO-8 (orange), PROGRS-4 (green), and PROGRS-8 (blue). 
Shaded regions indicate standard deviation across evaluation runs.}
    \label{fig:pass_k_all}
\end{figure}

\noindent\textbf{Efficiency across rollout and token budgets.}
Fig.~\ref{fig:efficiency} summarizes the accuracy--compute trade-off under different budget views.
\textbf{Accuracy vs.\ rollout budget} (Fig.~\ref{fig:efficiency}a) shows that PROGRS-4 reaches accuracy comparable to
DAPO baselines with substantially fewer rollouts, while PROGRS-8 matches or exceeds DAPO-16 at similar budget levels,
indicating a more favorable accuracy--computation frontier. \textbf{Token efficiency} (Fig.~\ref{fig:efficiency}b),
together with Table~\ref{tab:results_summary}, suggests that PROGRS achieves higher accuracy per generated token on
average, and that gains are not explained solely by longer generations. \textbf{Aggregate metrics} (Fig.~\ref{fig:efficiency}c--d)
further highlight the budget dependence: PROGRS-4 is most effective in constrained regimes, whereas PROGRS-8 provides
additional gains primarily at mid-range budgets, consistent with diminishing returns at higher rollout counts.
\noindent\textbf{Pass@K Performance Analysis}
\label{sec:passk_analysis}
Pass@K results (Fig.~\ref{fig:pass_k_all}) show that PROGRS consistently outperforms DAPO across benchmarks and sampling budgets. 
In many cases, PROGRS-4 matches or exceeds DAPO-16, indicating better sample efficiency. Although all methods improve as $K$ increases, the gap between PROGRS and DAPO remains, suggesting gains come not only from better exploration or sample ranking, but also from stronger base policy quality. 
Variance is generally lower or comparable under PROGRS, indicating similar or improved stability. 
Gains are largest on moderate- and high-accuracy benchmarks such as MATH-500 and AMC, and smaller in low-accuracy settings like AIME.

\begin{figure}[t!]
\centering
\includegraphics[trim={2cm 0 1cm 0},width=0.90\columnwidth]{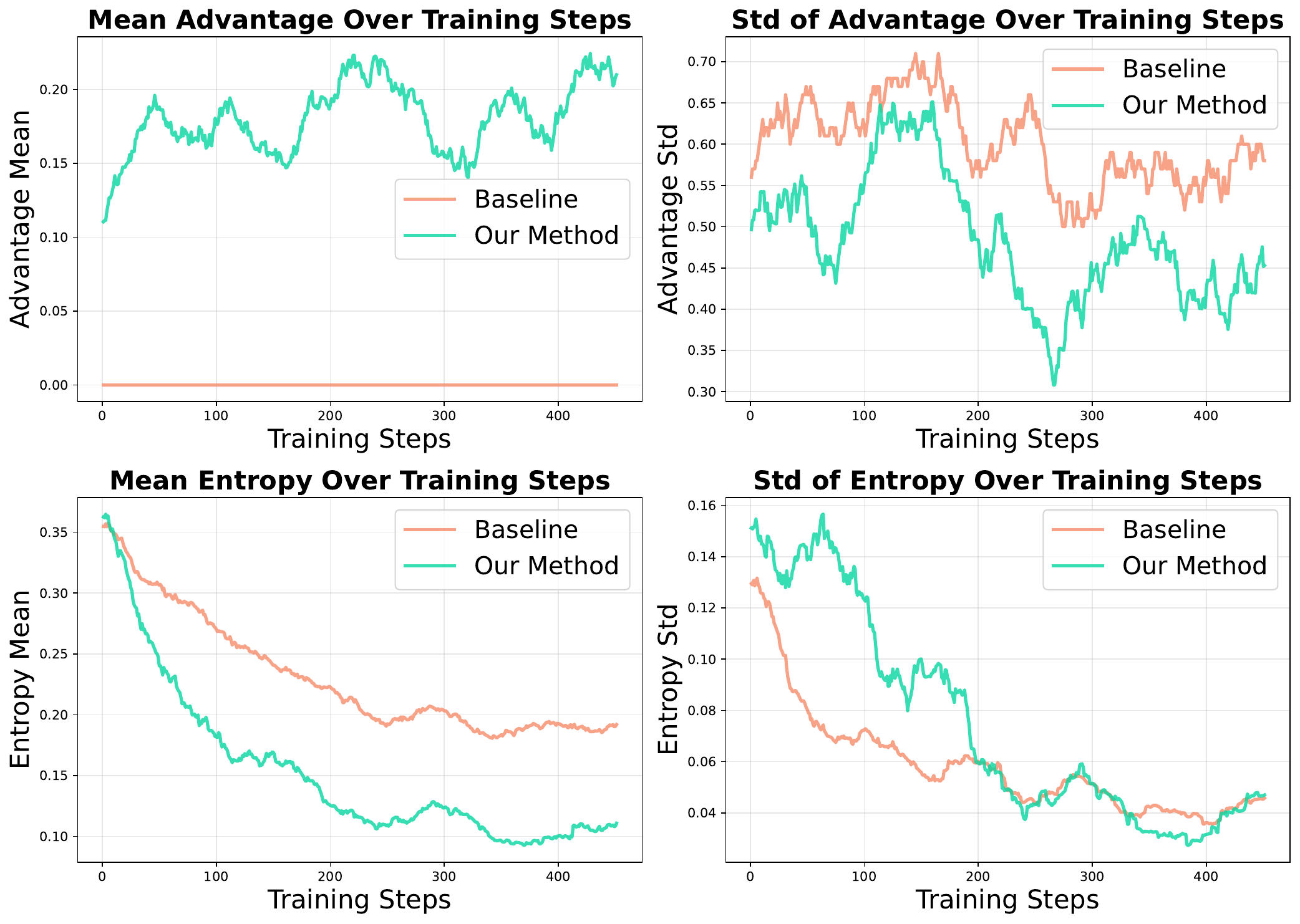}
\caption{Training dynamics of DAPO vs.\ PROGRS: mean and std.\ of advantages and per-token entropy over training (smoothed with a 50-step moving average).}
\label{fig:advantage-entropy}
\end{figure}

\noindent\textbf{Training Dynamics Analysis.}
We analyze optimization behavior by tracking advantage statistics and policy entropy during training (Fig.~\ref{fig:advantage-entropy}). Compared to DAPO, PROGRS exhibits more stable dynamics.
It maintains a more consistent advantage signal with lower variance, indicating reduced sensitivity to outlier trajectories and more stable gradient updates. In contrast, the baseline shows larger fluctuations.
Entropy under PROGRS decreases more smoothly with smaller variance, suggesting a more gradual and stable transition from exploration to exploitation. Overall, these results indicate improved optimization stability under PROGRS.

\subsection{Ablation Studies}
\label{sec:ablation_results}

PROGRS introduces two mechanisms designed to make process-reward guidance useful without undermining verifiable outcome supervision: (i) \emph{outcome-conditioned centering} (Eq.~2), which removes any systematic PRM-driven advantage offset on incorrect trajectories while preserving relative preferences within the incorrect set; and (ii) a \emph{coherence penalty}, which downweights trajectories whose step-level PRM scores exhibit abrupt local fluctuations. To isolate the contribution of each mechanism, we ablate them one at a time while holding fixed the policy model, PRM, rollout budget, decoding/evaluation protocol, and all other hyperparameters. 

\textbf{Effect of the coherence penalty ($\alpha_{\text{coh}}=0$).}
We first remove the coherence term while \emph{retaining} outcome-conditioned centering, to test whether coherence contributes beyond the core safeguard against PRM misalignment. Disabling coherence reduces accuracy relative to the full PROGRS configuration on most benchmarks, with particularly clear drops on AMC-2023 (from 59.00 to 50.50) and MATH-500 (from 74.92 to 71.82), and smaller but consistent decreases on MinervaMath and OlympiadBench. In parallel, average generation length typically decreases (e.g., MATH-500 from 683.74 to 669.70 tokens), indicating that coherence removal tends to shorten outputs but at the cost of correctness. This pattern is consistent with the intended role of coherence: it does not change which solutions are ultimately rewarded by verifiable outcomes, but it improves the \emph{quality} of process guidance by suppressing volatile step-level PRM signals that can otherwise encourage locally confident but unstable reasoning segments. The changes in run-to-run variability are mixed across datasets, but several tasks show increased variability when coherence is removed (e.g., AIME-24 and MATH-500), suggesting coherence can also act as a stabilizer for the PRM-derived component of the advantage.

\textbf{Effect of outcome-conditioned centering (No Centering).}
We next remove outcome-conditioned centering while keeping the PRM signal and all other settings unchanged, to directly test the failure mode that motivates PROGRS: absolute PRM scores may systematically over-reward trajectories that appear locally coherent even when the final answer is incorrect. This ablation produces the largest performance degradation overall, including a marked drop on MATH-500 (74.92 to 67.78) and a consistent decrease on OlympiadBench (35.06 to 30.77). Crucially, removing centering also tends to \emph{increase} generation length on harder or out-of-distribution settings (e.g., AIME-25 from 1156.79 to 1273.70 tokens; OlympiadBench from 958.38 to 975.54 tokens), indicating that without centering the model is incentivized to produce longer trajectories that score well under the PRM but do not translate into higher correctness. This directly matches the hypothesized misalignment: when incorrect samples can receive a net positive PRM offset, the learning signal can drift toward verbose, PRM-favored reasoning that competes with outcome supervision rather than complementing it.
Together, these ablations support a clear division of labor between the two components. Outcome-conditioned centering is the primary safeguard that preserves \emph{outcome dominance} by preventing systematic reward uplift on incorrect trajectories. 
Without it, PRM scores can drive reward hacking and reduce both accuracy and efficiency. The coherence penalty provides a secondary but meaningful benefit by refining the process signal within the safe regime enforced by centering: it tends to improve Pass@1 while also preventing pathological short/unstable reasoning patterns that arise when step-level PRM confidence is locally erratic. In addition, centering alone remains competitive with (and often stronger than) the outcome-only baselines. 
However, the full method achieves the most robust improvements across benchmarks when both mechanisms are combined.
% ===================== NEW MAIN-TEXT MATERIAL FROM APPENDIX =====================
\subsection{Comparison with Prior PRM-Based Methods}
\label{sec:prior_prm_main}

We compare \textsc{PROGRS} with recent PRM-based RL methods, PROF-GRPO~\cite{ye2025prof} and PRL~\cite{yao2026prl}, using a shared Qwen2.5-Math-1.5B-base model. For \textsc{PROGRS-8}, results are evaluated using Avg@8/16 metrics over rollouts.

Results for PROF-GRPO and PRL are taken from their original papers. Although all methods use the same base model, differences in training and optimization make the comparison indicative rather than strictly controlled.

\textsc{PROGRS} outperforms prior methods on AMC23, MATH-500, OlympiadBench, and AIME24, while remaining competitive on MinervaMath, showing robust gains under distribution shift.

\begin{table}[t]
\centering
\caption{Comparison with prior PRM-based methods under a shared base model. Avg@8/16 (\%).}
\label{tab:prof-prl}
\small
\setlength{\tabcolsep}{3pt}
\renewcommand{\arraystretch}{0.9}
\begin{tabular}{lcccc}
\toprule
Dataset &
\makecell{PROGRS\\Avg@8} &
\makecell{PROGRS\\Avg@16} &
\makecell{PROF-GRPO\\Avg@16} &
\makecell{PRL\\Avg@8} \\
\midrule
AMC23    & 58.75 & 60.47 & 49.1 & 45.94 \\
MATH-500 & 75.42 & 74.82 & 73.2 & 71.00 \\
Minerva  & 22.15 & 22.36 & 30.0 & 28.63 \\
Olympiad & 37.91 & 38.02 & 36.1 & 33.57 \\
AIME24   & 10.83 & 10.21 & 9.6  & 9.17 \\
\bottomrule
\end{tabular}
\end{table}

\noindent\textbf{Comparison with Alternative PRM Integrations.}
\label{sec:alt_prm_main}
Beyond these ablations, we compare \textsc{PROGRS}
against several alternative ways of incorporating process-level supervision into RLVR under
a matched rollout budget ($K=4$).
Specifically, we evaluate: \emph{Global Centering}, which subtracts a single group-wise
mean across all samples; \emph{Rank Shaping}, which converts PRM scores into pairwise
Bradley--Terry style logits; \emph{Step Shaping}, which applies per-step rewards without
coherence modulation; and \emph{PRM Filtering}, which discards low-coherence trajectories
before optimization.
As shown in Table~\ref{tab:baselines}, \textsc{PROGRS}-4 achieves the highest Pass@1
on all six benchmarks, outperforming these alternatives on both in-distribution
(MATH-500) and out-of-distribution settings (AMC/AIME, MinervaMath, OlympiadBench).
This supports the view that outcome-conditioned centering, combined with coherence-aware
aggregation, provides a more reliable way to use PRM signals than purely global
normalization, rank-based shaping, or heuristic trajectory filtering.

\begin{table}[t]
\centering
\caption{Rollout-matched baselines ($K=4$). Pass@1 (\%, higher better),
mean$\pm$std over 10 runs.}
\label{tab:baselines}
\footnotesize
\setlength{\tabcolsep}{2.5pt}
\renewcommand{\arraystretch}{1.1}
\resizebox{\columnwidth}{!}{%
\begin{tabular}{clllllll}
\toprule
 & Model & AIME24 & AIME25 & AMC23 & MATH-500 & Minerva & Olympiad \\
\midrule
\multirow{5}{*}{\rotatebox{90}{Pass@1 (\%)}}
& Global Centering  & 13.3$\pm$2.6 & 3.0$\pm$3.1 & 50.8$\pm$3.5 & 63.8$\pm$0.7 & 14.1$\pm$0.6 & 30.4$\pm$1.0 \\
& Rank Shaping      & 3.0$\pm$2.3  & 8.3$\pm$1.7 & 35.0$\pm$4.0 & 60.9$\pm$1.0 & 12.2$\pm$0.6 & 24.4$\pm$1.0 \\
& Step Shaping      & 5.7$\pm$3.4  & 4.3$\pm$3.7 & 44.0$\pm$3.0 & 65.7$\pm$0.9 & 18.3$\pm$0.7 & 28.9$\pm$0.8 \\
& PRM Filtering     & 6.7$\pm$2.6  & 3.0$\pm$1.8 & 47.0$\pm$4.2 & 63.0$\pm$0.7 & 15.7$\pm$0.7 & 27.4$\pm$1.1 \\
& \textbf{PROGRS-4} & 10.7$\pm$2.0 & 12.0$\pm$1.6 & 50.3$\pm$2.6 & 74.1$\pm$1.0 & 23.6$\pm$0.7 & 35.7$\pm$0.9 \\
\bottomrule
\end{tabular}}
\end{table}

% ------------------------------------------------------------------------------

\subsection{Additional Analysis}
\label{sec:additional_analysis}

\noindent\textbf{Scaling to Larger Policy Models.}
To assess whether the benefits of \textsc{PROGRS} persist at larger model scales, we
replicate our comparison with DAPO using Qwen2.5-Math-7B as the policy model while
keeping the PRM architecture and training protocol unchanged.
Table~\ref{tab:scaling-main} reports Pass@1 accuracy on all benchmarks for DAPO baselines
(8 and 16 rollouts) and \textsc{PROGRS} variants (4 and 8 rollouts).
Across datasets, \textsc{PROGRS}-8-7B matches or exceeds DAPO-16-7B while using only half
the rollout budget, and achieves the strongest performance on MATH-500, MinervaMath, and
OlympiadBench.
These results show that the proposed advantage construction remains effective at larger model scales and preserves a favorable accuracy--budget trade-off.

\begin{table}[t]
\centering
\caption{Scaling to a 7B policy. Pass@1 (\%), mean$\pm$std over 10 runs.}
\label{tab:scaling-main}
\footnotesize
\setlength{\tabcolsep}{2.5pt}
\renewcommand{\arraystretch}{1.1}
\resizebox{\columnwidth}{!}{%
\begin{tabular}{clllllll}
\toprule
 & Model & AIME24 & AIME25 & AMC23 & MATH-500 & Minerva & Olympiad \\
\midrule
\multirow{4}{*}{\rotatebox{90}{Pass@1 (\%)}}
& DAPO16-7B        & 24.0$\pm$1.3 & 13.3$\pm$2.1 & 66.0$\pm$4.1 & 77.3$\pm$0.5 & 36.0$\pm$0.9 & 38.8$\pm$0.9 \\
& DAPO8-7B         & 20.0$\pm$2.1 & 10.7$\pm$4.4 & 58.0$\pm$6.0 & 68.2$\pm$1.1 & 26.0$\pm$1.2 & 34.5$\pm$0.6 \\
& PROGRS4-7B       & 19.3$\pm$1.3 & 10.7$\pm$4.9 & 53.0$\pm$1.9 & 70.9$\pm$0.9 & 29.2$\pm$0.4 & 36.3$\pm$0.6 \\
& \textbf{PROGRS8-7B} & 22.0$\pm$2.7 & 12.7$\pm$2.5 & 57.0$\pm$1.9 & \textbf{81.5$\pm$0.5} & \textbf{37.5$\pm$0.5} & \textbf{41.6$\pm$0.6} \\
\bottomrule
\end{tabular}}
\end{table}

% ------------------------------------------------------------------------------

\begin{figure}[th!]
\centering
\includegraphics[width=\columnwidth]{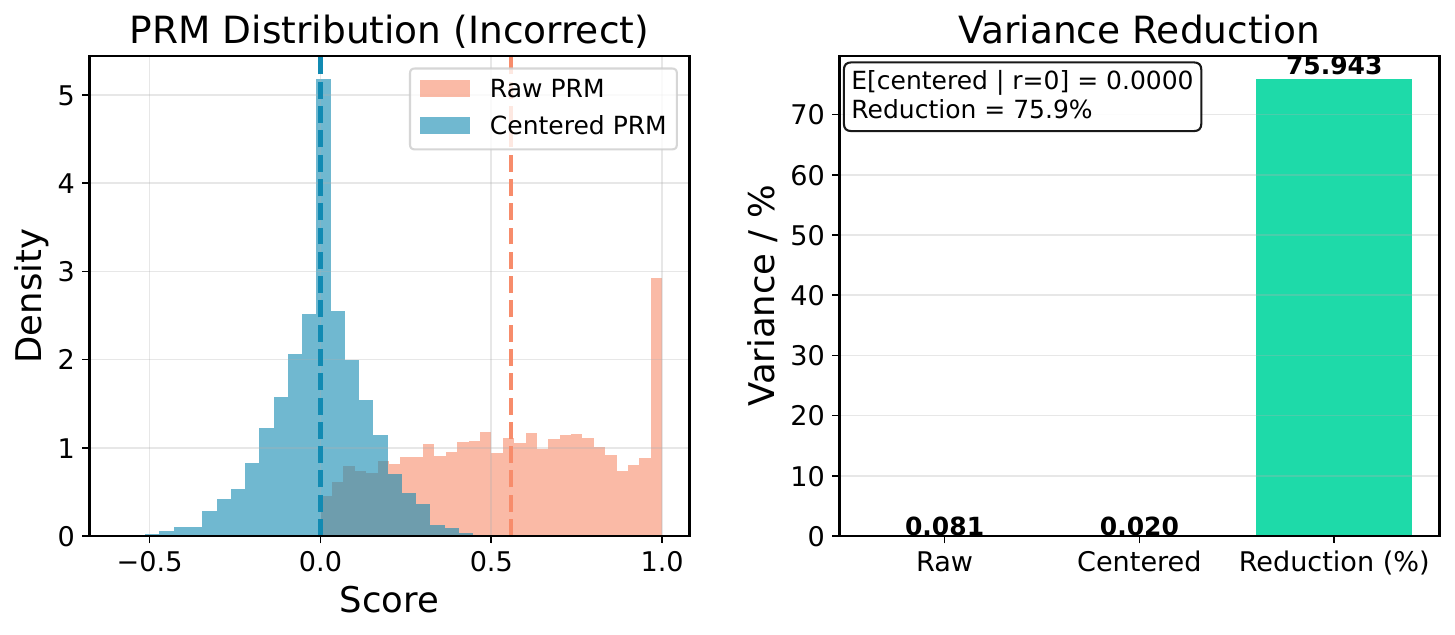}
\caption{Effect of outcome-conditioned centering on incorrect trajectories. 
Left: histogram of raw PRM scores vs. centered scores for $r=0$ samples. 
Right: variance comparison; centering reduces variance by $75.9\%$ while enforcing zero mean on incorrect bonuses ($\mathbb{E}[S^{\text{PRM}}|r=0]\approx0.558$, $\operatorname{Var}[S^{\text{PRM}}|r=0]\approx0.081$; $\mathbb{E}[\tilde{S}|r=0]=0$, $\operatorname{Var}[\tilde{S}|r=0]\approx0.020$).}
\label{fig:centering-var}
\end{figure}

\noindent\textbf{Variance Reduction from Outcome-Conditioned Centering.}
Finally, we highlight the variance-reduction effect of outcome-conditioned centering on
incorrect trajectories.
Fig.~\ref{fig:centering-var} compares the distribution of raw PRM scores against the
centered scores for samples with $r=0$, and quantifies their empirical variance.
Centering enforces a zero-mean constraint on incorrect bonuses and removes a dominant
shared noise mode, reducing the variance of PRM-derived contributions by roughly
three quarters.

% ---------------------- Conclusion --------------------
\section{Conclusion}
\label{sec:conclusion}
We introduced \textsc{PROGRS}, a simple method for incorporating process reward models (PRMs) into outcome-verifiable RL for mathematical reasoning. It addresses a central failure mode of PRM-based training: absolute process scores can be misaligned with final correctness and may reward fluent but incorrect trajectories. \textsc{PROGRS} combines outcome-conditioned centering, which removes systematic PRM bias on incorrect trajectories, with coherence-based weighting, which downweights unstable process signals. Because it adds no trainable components and uses a frozen PRM only for scoring, it integrates naturally into GRPO/DAPO-style training. Experiments across multiple math benchmarks show consistent Pass@1 gains and, in several settings, improved rollout efficiency, while ablations confirm that these gains come from centering and coherence rather than from simply adding process rewards.

Although \textsc{PROGRS} reduces sensitivity to absolute PRM miscalibration, it still relies on useful relative PRM preferences. Extending it beyond fully verifiable tasks will require alternative signals for defining outcome groups, such as consistency constraints, tool feedback, or learned evaluators. Overall, our results show that process rewards are most effective when used as outcome-constrained relative signals rather than absolute training targets.

\section*{Acknowledgment}
The authors gratefully acknowledge the computing time made available to them on the  high-performance computer Capella at the NHR Center at TU Dresden (ZIH). This center is jointly supported by the Federal Ministry of Research, Technology and Space and the state governments participating in the National High-Performance Computing (NHR) joint funding program (https://www.nhr-verein.de/en/our-partners).

\bibliographystyle{IEEEtran}
\bibliography{references}
\end{document}